\documentclass{article}

\PassOptionsToPackage{numbers, compress}{natbib}
\usepackage[preprint]{neurips_2026}

\usepackage[utf8]{inputenc} 
\usepackage[T1]{fontenc}    
\usepackage{hyperref}       
\usepackage{url}            
\usepackage{booktabs}       
\usepackage{amsfonts}       
\usepackage{nicefrac}       
\usepackage{microtype}      
\usepackage{xcolor}         
\usepackage{multirow}
\usepackage{graphicx}
\usepackage{amsmath}
\usepackage{makecell}
\usepackage{pifont}
\usepackage{array}
\usepackage{placeins}
\usepackage{subcaption}
\newcommand{\cmark}{\ding{51}}
\newcommand{\xmark}{\ding{55}}
\newcolumntype{C}[1]{>{\centering\arraybackslash}m{#1}}

\title{When Noise Meets Long-Tail: Feature-Threshold Dual Calibration for Robust Pseudo-Labeling}

\author{%
  Ping Guo\textsuperscript{1,2}
  \qquad
  Zhiqi Huang\textsuperscript{1,3}\thanks{Corresponding author.}
  \qquad
  Xinran Li\textsuperscript{1,2}
  \\[4pt]
  \textsuperscript{1}Graduate School of Information, Production and Systems,\\
  Waseda University, Kitakyushu, Japan
  \\[2pt]
  \textsuperscript{2}School of Software, Dalian University of Technology, Dalian, China
  \\[2pt]
  \textsuperscript{3}School of Artificial Intelligence,\\
  The Chinese University of Hong Kong, Shenzhen, China
  \\[4pt]
  \href{mailto:sr782442143@akane.waseda.jp}
       {\texttt{sr782442143@akane.waseda.jp}}\\
  \href{mailto:zhiqi.huang@akane.waseda.jp}
       {\texttt{zhiqi.huang@akane.waseda.jp}}
  \quad
  \href{mailto:lixinran@ruri.waseda.jp}
       {\texttt{lixinran@ruri.waseda.jp}}
}

\begin{document}
\raggedbottom
\setlength{\textfloatsep}{6pt plus 1pt minus 1pt}

\maketitle

\begin{abstract}
Pseudo-labeling has become a cornerstone of learning from unlabeled data in semantic segmentation. Yet its effectiveness drops sharply in real-world scenarios where strong imaging noise and long-tailed class distributions occur together. We trace this failure to a vicious cycle of pseudo-label degradation. Imaging noise entangles foreground and background features, lowering prediction confidence across all classes, while long-tailed distributions leave tail classes with far fewer training samples and inherently lower confidence. Under fixed high-threshold filtering, these tail-class predictions are systematically filtered out, so they receive no supervision from unlabeled data and thus features keep degrading in subsequent iterations. Critically, noise and long-tail are not independent obstacles but mutually amplifying ones, and addressing either alone is insufficient. To break this cycle, we propose FTC-Seg, a Feature-Threshold dual-Calibration framework built on a standard teacher-student framework. At the feature level, Orthogonal Prototype Reconstruction (OPR) uses a set of learnable orthogonal prototypes to residually purify pixel-wise features, widening the margin between weak foreground targets and noisy backgrounds. At the threshold level, Adaptive Threshold Calibration (ATC) dynamically adjusts class-specific thresholds based on learning difficulty and prediction-distribution bias, rescuing low-confidence pseudo-labels of tail classes from systematic exclusion. Extensive experiments on four public benchmarks spanning three distinct noise modalities show that FTC-Seg achieves strong performance against state-of-the-art methods, with particularly substantial gains on tail classes. Our results establish that jointly calibrating features and thresholds is essential for robust pseudo-labeling under compounded noise and class imbalance.
\end{abstract}

\vspace{-12pt}

\begin{center}
The code is available at \url{https://github.com/pingggg516/FTC-Seg}.
\end{center}

\section{Introduction}

Pseudo-labeling, where a model generates supervision from its own confident predictions on unlabeled data, has become a cornerstone for semantic segmentation tasks under limited annotation. Methods such as FixMatch~\citep{sohn2020fixmatch}, UniMatch~\citep{yang2023revisiting}, and SemiVL~\citep{hoyer2024semivl} have achieved remarkable performance on standard benchmarks like Pascal VOC~\citep{everingham2010pascal} and Cityscapes~\citep{cordts2016cityscapes}, all relying on a common assumption: that confident predictions filtered by a sufficiently high threshold could provide reliable supervision signals. This assumption breaks down in real-world scenarios where the data distribution departs sharply from clean and balanced benchmarks.

A particularly challenging regime arises when strong imaging noise and long-tailed class distributions occur together, as is common in sonar perception, underwater optical imaging, and adverse-weather driving. Environmental interference and long-tailed class distributions are also documented in the MMIO industrial defect benchmark~\citep{zhang2025zero}. Prior research has progressed along two largely separate directions: noise-robust pseudo-labeling~\citep{zhang2025noise,scherer2022pseudo,tarvainen2017mean,li2026rsod,feng2026semi}, which targets unreliable predictions caused by image-level noise or label corruption; and long-tailed semi-supervised learning~\citep{wei2021crest, oh2022daso,wang2022freematch}, which addresses class imbalance. However, the specific regime in which the two factors interact within the pseudo-labeling pipeline of semantic segmentation has received little explicit attention, and produces a distinct failure mode that neither line alone resolves.

\begin{figure}[t]
  \centering
  \includegraphics[width=\linewidth]{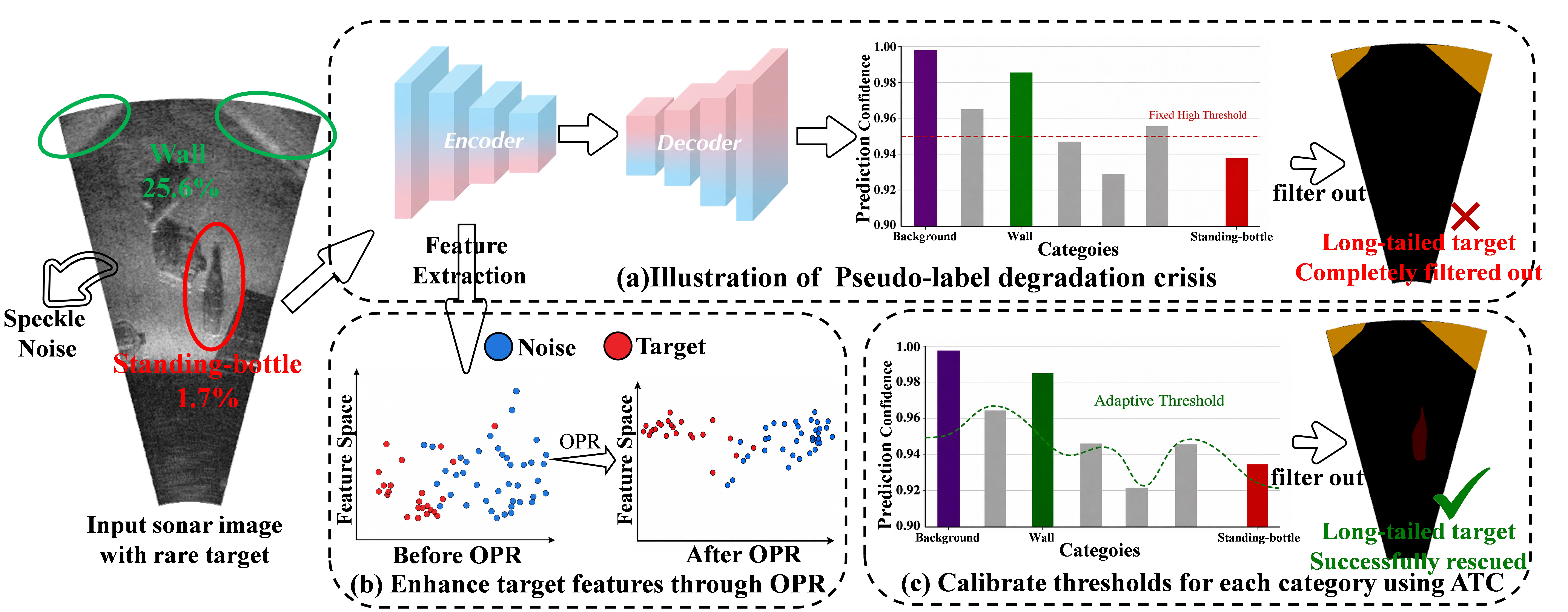}
  \caption{Illustration of the pseudo-label degradation crisis and our FTC-Seg motivation. (a) Traditional fixed high-threshold strategies are severely disturbed by speckle noise, systematically filtering out low-confidence predictions of long-tailed targets such as standing bottle. (b) At the feature level, the Orthogonal Prototype Reconstruction (OPR) enlarges the margin between target and background noise in the feature space. (c) At the threshold level, the Adaptive Threshold Calibration (ATC) dynamically lowers the threshold for rare classes, successfully rescuing long-tailed targets.}
  \label{fig:Motivation}

\end{figure}

\begin{figure}[t]
    \centering
    \hspace*{-0.22in}
    \includegraphics[width=\linewidth]{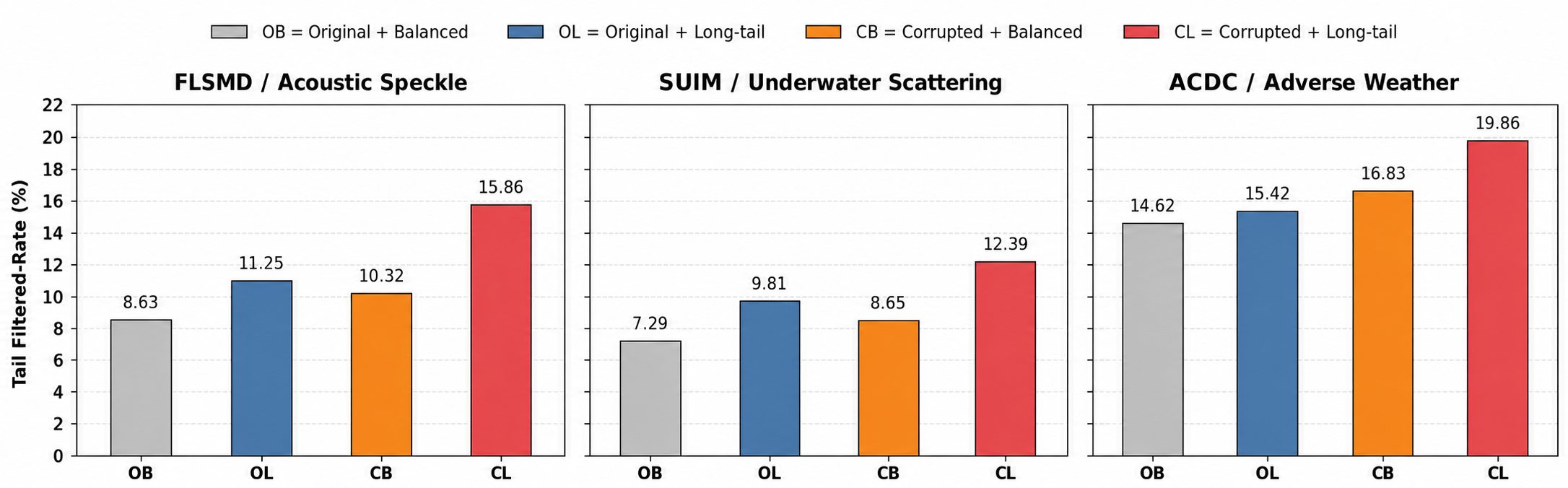}  
    \caption{Controlled decoupling analysis of tail-class pseudo-label filtering. OB/OL use original images and CB/CL apply domain-aligned synthetic corruption (multiplicative speckle for FLSMD, underwater scattering for SUIM, fog for ACDC) to the same image lists, while balanced subsets are obtained by sub-sampling head classes to match tail-class frequencies; masks are unchanged across conditions. The highest filtered-rate under CL across all three modalities indicates that corruption and long-tail jointly aggravate tail-class exclusion.}
\label{fig:Circle}
\end{figure}

We trace this failure to a vicious cycle of pseudo-label degradation, as illustrated in Figure~\ref{fig:Motivation}. Imaging noise blurs foreground objects into the surrounding background, lowering prediction confidence across all classes, while long-tailed distributions leave tail classes with even lower confidence due to weaker representations. Under a fixed high-confidence threshold, these tail predictions are filtered out almost entirely, so tail classes receive no supervision from unlabeled data and their features keep degrading. The crucial point is that noise and long-tail are not independent obstacles but mutually amplifying ones: noise creates the confidence gap, long-tail decides who falls into it, and fixed thresholds turn the gap into permanent exclusion. Figure~\ref{fig:Circle} verifies this super-linear interaction through a controlled 2×2 decoupling experiment: across FLSMD, SUIM and ACDC datasets, the tail filtered-rate under joint corruption and long-tail (CL) exceeds the linear sum of each factor alone by 1.2–2.9 percentage points. The four conditions differ in exactly one factor at a time: OB/CB share an identical balanced image list (head classes sub-sampled to match tail-class frequency) while OL/CL share an identical long-tail list, with the same pixel-level masks reused across all four cells. For ACDC, the clean Cityscapes counterpart of each image serves as the original version, providing a noise-free baseline that is unavailable for the sonar and underwater modalities.

To break this cycle, we propose FTC-Seg, which calibrates pseudo-labeling at both ends of the failure chain through Orthogonal Prototype Reconstruction (OPR) at the feature level (Figure~\ref{fig:Motivation}(b)) and Adaptive Threshold Calibration (ATC) at the threshold level (Figure~\ref{fig:Motivation}(c)). OPR restructures pixel features through orthogonal prototypes to make tail predictions more confident, while ATC replaces the fixed threshold with class-specific ones that relax otherwise-rejected tail predictions, together breaking the degradation cycle.

Our main contributions can be summarised as follows: \textbf{(i) A failure mechanism:} we identify and characterize a vicious cycle of pseudo-label degradation that arises from the mutually amplifying effects of imaging noise and long-tailed distributions. \textbf{(ii) A general principle:} we argue that addressing this cycle requires joint calibration at both the feature level and the threshold level, and instantiate this principle as FTC-Seg, a framework grounded in a teacher-student pseudo-labeling pipeline. \textbf{(iii) Empirical evidence across noise modalities:} through experiments on four public benchmarks covering three distinct types of imaging noise—acoustic speckle, underwater optical scattering and adverse-weather imaging—we show that the same failure mode appears under all three noise types, and that joint feature-threshold calibration consistently resolves it.

\section{Related Work}
\paragraph{Semi-supervised semantic segmentation.}
Modern semi-supervised semantic segmentation builds on the teacher-student paradigm, where Mean Teacher~\citep{tarvainen2017mean} stabilizes pseudo-label generation through an exponential moving average of student weights. FixMatch~\citep{sohn2020fixmatch} extends this line by introducing weak-to-strong consistency, in which pseudo-labels generated from weakly-augmented inputs are filtered by a fixed high-confidence threshold and then supervise predictions on strongly-augmented ones. Subsequent work introduces cross-network supervision~\citep{chen2021semi} and and multi-level perturbations~\citep{yang2023revisiting} for segmentation, while DTCG~\citep{liu2024two} uses complementary dual-teacher guidance for semi-supervised elliptical object detection. UniMatch V2~\citep{yang2025unimatch} further leverages DINOv2 as a strong backbone, while SemiVL~\citep{hoyer2024semivl} and Beyond-Pixels~\citep{howlader2024beyond} introduce vision-language guidance and multi-scale patch-based supervision respectively. Despite their success on clean and balanced benchmarks such as Pascal VOC and Cityscapes, the fixed-threshold filtering shared by most of these methods is class-agnostic and, in noisy and long-tailed regimes, disproportionately eliminates tail-class predictions whose confidence is suppressed by both noise and limited training data.

\paragraph{Long-tailed semi-supervised learning.}
A separate line of work targets the imbalance bias of pseudo-labeling under long-tailed distributions, branching into two families. Distribution-level methods such as CReST~\citep{wei2021crest} and DASO~\citep{oh2022daso} for classification re-balance training by aligning pseudo-label distributions with the labeled prior or re-sampling rare classes more aggressively, but do not modify the threshold itself, so tail predictions can still be eliminated at the generation stage before any re-balancing takes effect. Threshold-level methods such as FlexMatch~\citep{zhang2021flexmatch} and FreeMatch~\citep{wang2022freematch}, though originally proposed for classification, replace the single global threshold with class-specific ones derived from the model's own learning state on unlabeled data; however, this self-referential signal cannot distinguish genuine improvement from over-prediction bias, and without an external signal to the labeled distribution, threshold adaptation can reinforce confirmation bias. More fundamentally, both families presuppose that tail-class predictions are at least separable in feature space once passed the filter—an assumption that breaks under strong imaging noise where tail features are actively confused with the background.

\paragraph{Noise-robust pseudo-labeling.}
Another line of work addresses low-quality pseudo-labels under noise through two strategies. Temporal-ensembling methods such as Mean Teacher~\citep{tarvainen2017mean} smooth short-term prediction noise via exponential moving averages, while augmentation and consistency based methods—including pseudo-label noise suppression~\citep{scherer2022pseudo} via cow-mix augmentation and noise-robust consistency regularization~\citep{zhang2025noise} combining multiple denoising perspectives—improve robustness through perturbation-invariant training. However, these methods operate on pseudo-labels that pass the confidence threshold; they offer no mechanism for cases where strong imaging noise pushes tail-class predictions below the threshold and discards them before denoising can take effect, leaving tail classes systematically excluded for being low-confidence rather than for being wrong.

\paragraph{Our position.}
The three lines above each address either noise or long-tail in isolation, but the separation breaks down when the two factors occur together: threshold or distribution-level re-balancing cannot rescue predictions whose features have already been corrupted by noise, while noise-robust techniques cannot recover tail predictions filtered out before they reach denoising. Resolving this coupled failure requires intervention at both the feature level, where noise and long-tail jointly degrade representations, and the threshold level, where the resulting confidence collapse is turned into permanent exclusion.

\begin{figure*}[t]
    \centering
    \includegraphics[width=\linewidth,height=0.36\textheight]{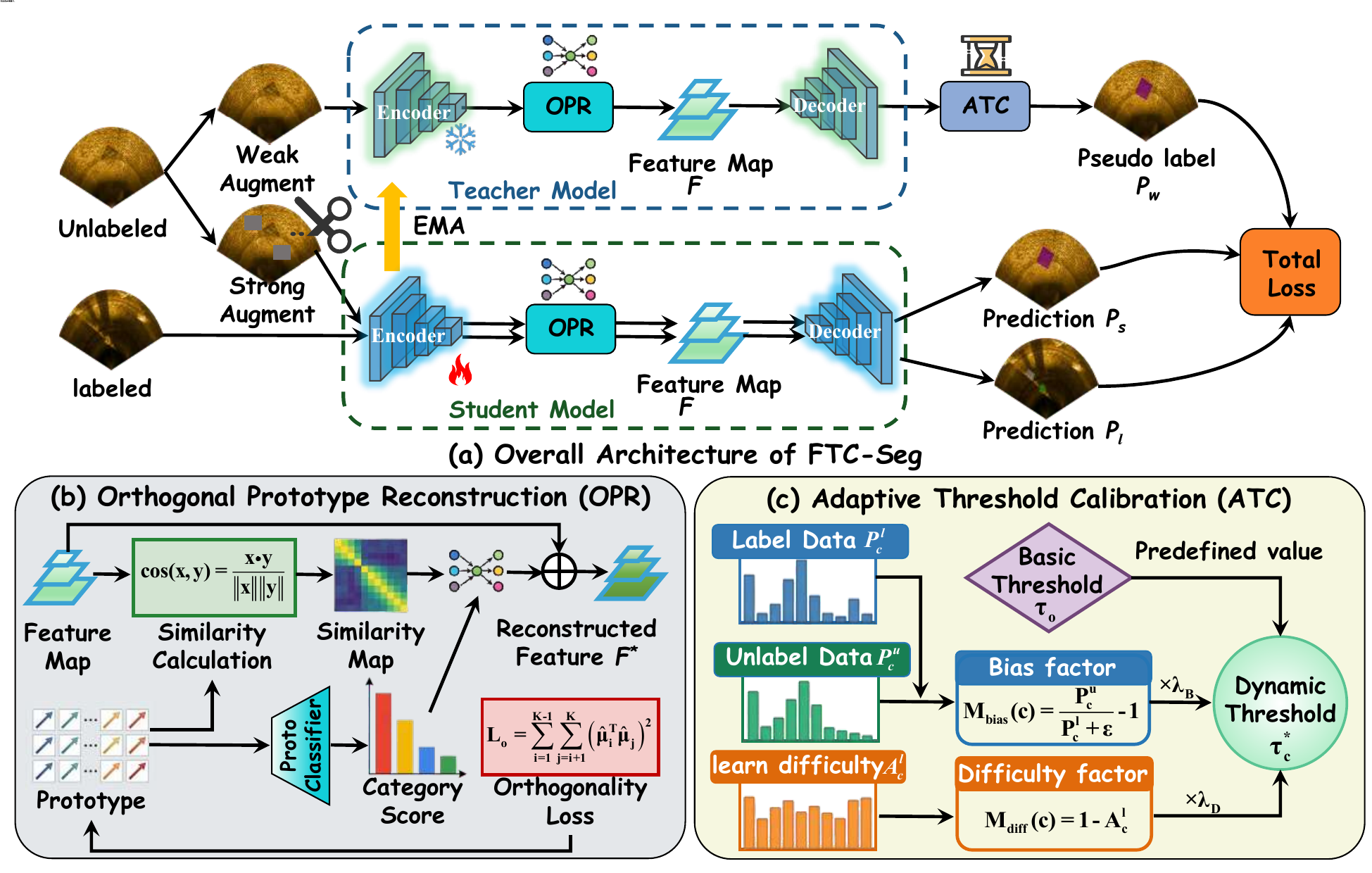}
    \caption{Overview of FTC-Seg framework. (a) Teacher-student architecture with OPR inserted between encoder and decoder, and ATC applied before pseudo-label filtering. (b) OPR computes similarities between features and learnable orthogonal prototypes, then aggregates foreground prototypes via a residual connection. (c) ATC combines a difficulty factor $\mathcal{M}_{\text{diff}}$ and a bias factor $\mathcal{M}_{\text{bias}}$ to produce class-specific thresholds.}
    \label{fig:workflow}
\end{figure*}

\section{Methodology}
To break the degradation cycle identified in Sec~1, we propose FTC-Seg, a feature–threshold dual-calibration framework introducing two modules: Orthogonal Prototype Reconstruction (OPR) at the feature level, which purifies pixel-wise features against noise-induced entanglement, and Adaptive Threshold Calibration (ATC) at the threshold level, which replaces the class-agnostic fixed threshold with class-specific ones to prevent low-confidence tail predictions from being systematically excluded.

\paragraph{Overall Framework.}
The overall architecture of FTC-Seg is illustrated in Figure~\ref{fig:workflow}(a). At each training iteration, the student receives a labeled image and a strongly-augmented view of an unlabeled image, while the teacher receives the weakly-augmented view of the same unlabeled image. Both branches share an encoder–decoder, with OPR inserted between encoder and decoder to refine the feature map $F$. The teacher is updated as an exponential moving average (EMA) of the student and provides pseudo-labels supervising the student's predictions on the strongly-augmented view.

\begin{table*}[t]   
  \centering
  \caption{Quantitative comparison of mIoU (\%) on the FLSMD and FSSG datasets. The best results are shown in \textbf{bold}, and the second best results are shown in \underline{\textit{italic}}. Results of FTC-Seg and all comparative methods are averaged over 3 random seeds (±std).} 
  \label{tab:FSSG_FLSMD_COMP}

  \small
  \setlength{\tabcolsep}{7pt}
  \renewcommand{\arraystretch}{0.95}

  \resizebox{\textwidth}{!}{
  \begin{tabular}{c|c|c|ccc|ccc}  
    \toprule 
    \multirow[c]{2}{*}{\textbf{Method}} & 
    \multirow[c]{2}{*}{\textbf{Venue}} & 
    \multirow[c]{2}{*}{\textbf{Encoder}} & 
    \multicolumn{3}{c|}{\textbf{FLSMD}} & 
    \multicolumn{3}{c}{\textbf{FSSG}}\\
    & & & 2\% & 5\% & 10\% & 2\% & 5\% & 10\%\\
    \midrule

    Labeled Only & -- & RN-101    
          & 46.47\,{\scriptsize $\pm$1.17} & 55.15\,{\scriptsize $\pm$0.47} & 61.94\,{\scriptsize $\pm$0.83} & 17.64\,{\scriptsize $\pm$1.35} & 32.23\,{\scriptsize $\pm$0.66} & 52.27\,{\scriptsize $\pm$1.01}\\

    Labeled Only & -- & DINOv2-S    
          & 51.08\,{\scriptsize $\pm$0.50} & 61.02\,{\scriptsize $\pm$0.95} & 68.48\,{\scriptsize $\pm$0.70} & 35.61\,{\scriptsize $\pm$0.89} & 41.67\,{\scriptsize $\pm$1.42} & 54.58\,{\scriptsize $\pm$0.32}\\

    \midrule 

    AEL~\cite{hu2021semi} & NIPS'21 & RN-101  
          & 52.70\,{\scriptsize $\pm$1.54} & 64.89\,{\scriptsize $\pm$1.24} & \underline{\textit{70.73}}\,{\scriptsize $\pm$0.88} & 53.51\,{\scriptsize $\pm$1.02} & 57.84\,{\scriptsize $\pm$0.98} & 63.08\,{\scriptsize $\pm$0.46}\\    

    Dual Teacher~\cite{na2023switching} & NIPS'23 & MiT-B2  
          & \underline{\textit{56.23}}\,{\scriptsize $\pm$0.70} & \underline{\textit{66.09}}\,{\scriptsize $\pm$0.56} & 69.61\,{\scriptsize $\pm$0.71} & 58.31\,{\scriptsize $\pm$0.80} & 60.16\,{\scriptsize $\pm$0.93} & \underline{\textit{65.91}}\,{\scriptsize $\pm$1.01}\\  

    UniMatch~\cite{yang2023revisiting} & CVPR'23 & RN-101  
          & 50.70\,{\scriptsize $\pm$1.47} & 56.14\,{\scriptsize $\pm$0.68} & 61.66\,{\scriptsize $\pm$1.34} & 58.39\,{\scriptsize $\pm$0.88} & 60.87\,{\scriptsize $\pm$0.70} & 63.15\,{\scriptsize $\pm$1.27}\\  

    Beyond-Pixels~\cite{howlader2024beyond} & ECCV'24 & RN-101  
          & 48.19\,{\scriptsize $\pm$0.60} & 55.52\,{\scriptsize $\pm$1.32} & 60.49\,{\scriptsize $\pm$0.93} & 57.20\,{\scriptsize $\pm$1.08} & \underline{\textit{62.37}}\,{\scriptsize $\pm$0.88} & 64.52\,{\scriptsize $\pm$0.36}\\  

    SemiVL~\cite{hoyer2024semivl} & ECCV'24 & RN-101  
          & 53.38\,{\scriptsize $\pm$1.16} & 65.16\,{\scriptsize $\pm$0.89} & 70.02\,{\scriptsize $\pm$0.49} & \underline{\textit{58.85}}\,{\scriptsize $\pm$0.70} & 62.19\,{\scriptsize $\pm$0.29} & 65.24\,{\scriptsize $\pm$0.97}\\

    VC~\cite{chen2024virtual} & TPAMI'24 & RN-101  
          & 50.53\,{\scriptsize $\pm$1.63} & 59.36\,{\scriptsize $\pm$1.31} & 66.86\,{\scriptsize $\pm$1.34} & 49.76\,{\scriptsize $\pm$1.20} & 54.75\,{\scriptsize $\pm$0.88} & 59.27\,{\scriptsize $\pm$1.09}\\

    FTC-Seg (Ours) & -- & RN-101  
          & \textbf{58.33\,{\scriptsize $\pm$0.94}} & \textbf{67.20\,{\scriptsize $\pm$0.77}} & \textbf{71.01\,{\scriptsize $\pm$0.45}} 
          & \textbf{60.49\,{\scriptsize $\pm$0.82}} & \textbf{64.35\,{\scriptsize $\pm$0.61}} & \textbf{70.13\,{\scriptsize $\pm$1.14}}\\

    \midrule

    UniMatch V2~\cite{yang2025unimatch} & TPAMI'25 & DINOv2-S  
          & 57.24\,{\scriptsize $\pm$0.49} & 66.49\,{\scriptsize $\pm$0.51} & 69.81\,{\scriptsize $\pm$0.57} & 58.78\,{\scriptsize $\pm$1.32} & 61.21\,{\scriptsize $\pm$0.72} & 64.41\,{\scriptsize $\pm$0.75}\\

    FTC-Seg (Ours) & -- & DINOv2-S  
          & \textbf{62.55\,{\scriptsize $\pm$1.22}} & \textbf{68.91\,{\scriptsize $\pm$0.69}} & \textbf{72.54\,{\scriptsize $\pm$0.72}} 
          & \textbf{63.20\,{\scriptsize $\pm$0.98}} & \textbf{65.49\,{\scriptsize $\pm$1.07}} & \textbf{67.53\,{\scriptsize $\pm$0.68}}\\

    \bottomrule  
  \end{tabular}
  }

\end{table*}

\begin{figure}[t]
    \centering
    \includegraphics[width=\linewidth]{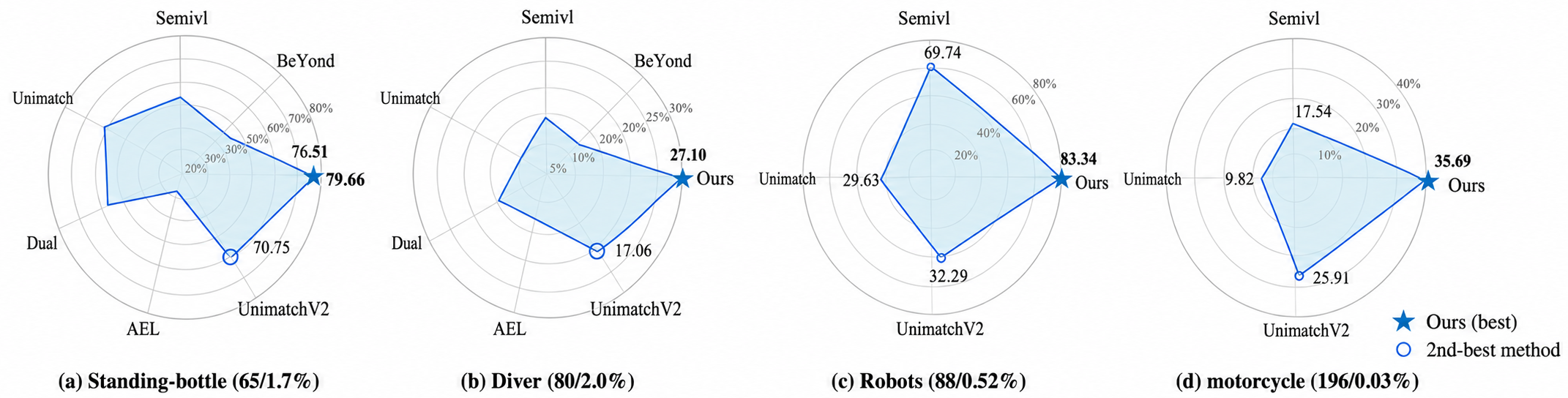}  
    \caption{Performance comparison on the rarest or second-rarest long-tailed categories from the FLSMD, FSSG, SUIM and ACDC datasets under the 2\% labeled-data setting.}
\label{fig:TailMiou}
\end{figure}

On the labeled data, the student is supervised by the standard pixel-wise cross-entropy loss:
\begin{equation}
\mathcal{L}_{sup}=-\frac{1}{HW}\sum_{i=1}^{H}\sum_{j=1}^{W}\sum_{c=1}^{C} Y_l^{(i,j,c)} \log P_l^{(i,j,c)},
\end{equation}
where $H$ and $W$ are the spatial dimensions, $C$ denotes the number of classes, $Y_l^{(i,j,c)}$ is the one-hot ground-truth and $P_l^{(i,j,c)}$ is the student's predicted probability for class $c$ at pixel $(i,j)$.

On the unlabeled data, the teacher's prediction $\tilde{P}_w$ on the weakly-augmented view is converted into a one-hot pseudo-label $P_w$ through class-specific confidence filtering:

\begin{equation}
P_w^{(i,j,c)}=
\begin{cases}
1, & c=\arg\max\limits_{c'}\tilde{P}_w^{(i,j,c')} \land M^{(i,j)}=1, \\
0, & \text{otherwise},
\end{cases}
\end{equation}

\begin{equation}
M^{(i,j)}=
\begin{cases}
1, & \max\limits_{c}\tilde{P}_w^{(i,j,c)}>\tau_c^{*}, \\
0, & \text{otherwise}.
\end{cases}
\end{equation}
The class-specific threshold $\tau_c^{*}$ is the central quantity threshold-level calibration acts upon; its construction is detailed in Sec~3.3. The pseudo-label $P_w$ then supervises the student's prediction $P_s$ on the strongly-augmented view through an unsupervised consistency loss:
\begin{equation}
\mathcal{L}_{unsup}=-\frac{1}{HW}\sum_{i=1}^{H}\sum_{j=1}^{W}\sum_{c=1}^{C} P_w^{(i,j,c)} \log P_s^{(i,j,c)}.
\end{equation}

The student minimizes $\mathcal{L}_{\text{sup}}$ and $\mathcal{L}_{\text{unsup}}$, while the teacher updates via EMA without backpropagation. OPR refines features before decoding; ATC sets $\tau_c^{*}$ before pseudo-label filtering.

\paragraph{Orthogonal Prototype Reconstruction (OPR).}
At the feature level, the degradation cycle manifests as noise-induced entanglement: under strong imaging noise, pixel-wise features of foreground targets are pulled toward the surrounding background, suppressing their prediction confidence, a feature-level intervention must therefore separate foreground from background before the decoder.  We argue that the intervention must satisfy three properties simultaneously: \textbf{(i)} anchoring each local pixel feature to a global signal stable against local noise; \textbf{(ii)} selectively admitting only the foreground portion, since mixing in background would re-introduce the entanglement; \textbf{(iii)} preserving the original feature rather than replacing it, since semantic segmentation depends on fine-grained spatial detail. OPR realizes these three properties through a small set of learnable vectors called \emph{prototypes}, shared across the dataset and converging to recurring semantic patterns. These global prototypes provide the stable signal required by property~(i).

As shown in Figure~\ref{fig:workflow}(b), we maintain $K$ learnable prototypes $P = \{\mu_1, \mu_2, \dots, \mu_K\} \in \mathbb{R}^{K \times C}$, where $C$ is the channel dimension of the encoder feature map $F \in \mathbb{R}^{C \times H \times W}$. Property~(i) is realized by measuring the cosine similarity between each pixel feature $F_{(h,w)}$ and each prototype $\mu_k$:
\begin{equation}
\label{eq:cosine_similarity}
S_{h,w}^{k} =
\frac{
F_{(h,w)}^{\top} \mu_{k}
}{
\left\| F_{(h,w)} \right\|_{2}
\left\| \mu_{k} \right\|_{2}
},
\end{equation}
where $S_{h,w}^{k}$ expresses how each locally noisy pixel aligns with each globally learned semantic pattern. Property~(ii) requires identifying which prototypes have converged to foreground semantics, obtained from the segmentation head whose weights encode class-level decision boundaries:
\begin{equation}
Z_c^{k} = W_c^{\top}\mu_k + b_c,
\label{eq:zck}
\end{equation}
where $W_c$ and $b_c$ are the segmentation head's weight and bias for class $c$. A suppression factor $\gamma_k$ then admits only prototypes whose dominant class lies in the foreground class set $\mathcal{C}_{fg}$:
\begin{equation}
\label{eq:Gamma}
\gamma_k = 
\begin{cases} 
1, & \arg\max_c Z_c^k \in C_{fg}, \\ 
0, & \text{otherwise} .
\end{cases},
\end{equation}

\begin{table*}[t]
  \centering
  \caption{Quantitative comparison of mIoU (\%) on the SUIM and ACDC datasets.The best results are shown in \textbf{bold}, and the second best results are shown in \underline{\textit{italic}}. Results of FTC-Seg and all comparative methods are averaged over 3 random seeds (±std).}
  \label{tab:suim_acdc_results}

  \small
  \setlength{\tabcolsep}{7pt}
  \renewcommand{\arraystretch}{0.95}

  \resizebox{\textwidth}{!}{
  \begin{tabular}{c|c|ccc|ccc}
    \toprule
    \multirow[c]{2}{*}{\textbf{Method}} &
    \multirow[c]{2}{*}{\textbf{Encoder}} &
    \multicolumn{3}{c|}{\textbf{SUIM}} &
    \multicolumn{3}{c}{\textbf{ACDC}} \\
    & & 2\% & 5\% & 10\% & 2\% & 5\% & 10\% \\
    \midrule

    Labeled Only & RN-101
          & 35.64\,{\scriptsize $\pm$1.44} & 45.47\,{\scriptsize $\pm$1.52} & 52.68\,{\scriptsize $\pm$0.97}
          & 36.62\,{\scriptsize $\pm$1.13} & 44.43\,{\scriptsize $\pm$0.91} & 48.47\,{\scriptsize $\pm$1.27} \\

    Labeled Only & DINOv2-S
          & 56.61\,{\scriptsize $\pm$0.80} & 62.56\,{\scriptsize $\pm$1.29} & 64.26\,{\scriptsize $\pm$1.15}
          & 52.60\,{\scriptsize $\pm$0.45} & 56.57\,{\scriptsize $\pm$0.62} & 66.98\,{\scriptsize $\pm$0.97} \\

    \midrule

    UniMatch & RN-101
          & 47.86\,{\scriptsize $\pm$2.17} & 49.17\,{\scriptsize $\pm$1.98} & 55.01\,{\scriptsize $\pm$1.71}
          & 45.09\,{\scriptsize $\pm$2.36} & 50.81\,{\scriptsize $\pm$1.49} & 51.66\,{\scriptsize $\pm$1.20}  \\

    SemiVL & RN-101
          & 55.99\,{\scriptsize $\pm$0.79} & 58.43\,{\scriptsize $\pm$0.60} & 60.15\,{\scriptsize $\pm$1.32}
          & 45.13\,{\scriptsize $\pm$0.57} & 50.65\,{\scriptsize $\pm$0.71}& 52.97\,{\scriptsize $\pm$0.97} \\

    FTC-Seg (Ours) & RN-101
          & 54.71\,{\scriptsize $\pm$0.53} & 59.05\,{\scriptsize $\pm$1.46} & 62.85\,{\scriptsize $\pm$0.97}
          & 45.56\,{\scriptsize $\pm$1.35} & 51.80\,{\scriptsize $\pm$0.89} & 55.73\,{\scriptsize $\pm$1.04}  \\

    \midrule

    UniMatch V2 & DINOv2-S
          &\underline{\textit{58.37}}\,{\scriptsize $\pm$0.92} & \underline{\textit{66.12}}\,{\scriptsize $\pm$1.07} & \underline{\textit{68.43}}\,{\scriptsize $\pm$0.85}
          & \underline{\textit{59.59}}\,{\scriptsize $\pm$2.48} & \underline{\textit{67.50}}\,{\scriptsize $\pm$1.63} & \underline{\textit{68.55}}\,{\scriptsize $\pm$1.77} \\

    FTC-Seg (Ours) & DINOv2-S
          & \textbf{66.91\,{\scriptsize $\pm$1.35}} & \textbf{69.86\,{\scriptsize $\pm$0.98}} & \textbf{70.21\,{\scriptsize $\pm$0.90}}
          & \textbf{61.68}\,{\scriptsize $\pm$1.42} & \textbf{69.82}\,{\scriptsize $\pm$0.97} & \textbf{70.09}\,{\scriptsize $\pm$0.86} \\

    \bottomrule
  \end{tabular}
  }
\end{table*}

\begin{table}[t]
  \centering
  \caption{Ablation studies of the proposed OPR and ATC components.}
  \label{tab:ablation_component_OPR_ATC}
  \scriptsize
  \setlength{\tabcolsep}{1.5pt}
  \renewcommand{\arraystretch}{1.06}

  \begin{minipage}[t]{0.49\linewidth}
    \vspace{0pt}
    \centering
    \textbf{(a) Ablation study of OPR components.}\par
    \vspace{2pt}

    \begin{tabular*}{\linewidth}{@{\extracolsep{\fill}} l|C{0.35cm}C{0.35cm}C{0.35cm}|cc|c @{}}
      \toprule
      \multirow{2}{*}{\textbf{Method}} 
      & \multicolumn{3}{c|}{\textbf{Component}} 
      & \multicolumn{2}{c|}{\textbf{Pseudo-Label Acc. (\%)}} 
      & \textbf{Global} \\
      \cmidrule(lr){2-4} \cmidrule(lr){5-6} \cmidrule(lr){7-7}
      & \textbf{Res.} & \textbf{$\mathcal{L}_o$} & \textbf{$r_k$}
      & \textbf{Global} & \textbf{Tail} 
      & \textbf{mIoU} \\
      \midrule
      Baseline            
      & -- & -- & --     
      & 67.03 & 56.56 & 58.75 \\
      w/o Residual        
      & \xmark & \cmark & \cmark 
      & 68.69 & \textbf{89.61} & 56.93 \\
      w/o $\mathcal{L}_o$ 
      & \cmark & \xmark & \cmark 
      & 71.30 & 81.04 & 59.26 \\
      w/o $r_k$           
      & \cmark & \cmark & \xmark 
      & 64.32 & 83.16 & 59.84 \\
      OPR                 
      & \cmark & \cmark & \cmark 
      & \textbf{72.33} & 87.80 & \textbf{62.55} \\
      \bottomrule
    \end{tabular*}
  \end{minipage}
  \hfill
  \begin{minipage}[t]{0.49\linewidth}
    \vspace{0pt}
    \centering
    \textbf{(b) Quantitative analysis of ATC components.}\par
    \vspace{2pt}

    \begingroup
    \renewcommand{\arraystretch}{1.23}
    \begin{tabular*}{\linewidth}{@{\extracolsep{\fill}} l|ccc|c|c @{}}
      \toprule
      \multirow{2}{*}{\textbf{Method}} 
      & \multicolumn{3}{c|}{\textbf{Tail Classes}} 
      & \textbf{Head Classes} 
      & \textbf{Global} \\
      \cmidrule(lr){2-4} \cmidrule(lr){5-5} \cmidrule(lr){6-6}
      & \textbf{Recall} & \textbf{Precision} & \textbf{IoU} 
      & \textbf{IoU} 
      & \textbf{IoU} \\
      \midrule
      Baseline 
      & 62.50 & 77.62 & 49.97 
      & 72.77 
      & 58.75 \\
      + $M_{\mathrm{diff}}$ 
      & 77.11 & 79.20 & 62.61 
      & 73.54 
      & 60.42 \\
      + $M_{\mathrm{bias}}$ 
      & 76.29 & 79.67 & 62.04 
      & 73.01 
      & 59.79 \\
      ATC 
      & \textbf{86.40} & \textbf{84.00} & \textbf{73.87} 
      & \textbf{75.37} 
      & \textbf{61.90} \\
      \bottomrule
    \end{tabular*}
    \endgroup
  \end{minipage}
\end{table}

Without this gating, background-aligned prototypes would contaminate the reconstruction, reinstating the entanglement OPR aims to suppress. Property~(iii) is realized through a residual aggregation:
\begin{equation}
\label{eq:aggregation}
F_{(h,w)}^{*}
=
F_{(h,w)}
+
\sum_{k=1}^{K}
\left(
\gamma_{k} \cdot S_{h,w}^{k}
\right)
\mu_{k}.
\end{equation}
The original feature $F_{(h,w)}$ is kept in full, while the prototype aggregation acts as a corrective term strengthening foreground-aligned components without overwriting spatial detail.

However, the three properties presuppose that the prototypes are distinct. If multiple prototypes collapse to similar directions, they produce nearly identical responses in Eq.~\ref{eq:cosine_similarity} and contribute redundantly to Eq.~\ref{eq:aggregation}, causing the prototype set to degenerate into a few repeated patterns rather than a diverse collection. We therefore impose an orthogonality regularizer on the prototype matrix:
\begin{equation}
\mathcal{L}_{O} = \sum_{i=1}^{K-1} \sum_{j=i+1}^{K} \left(\hat{\mu}_i^{\top}\hat{\mu}_j\right)^2,
\label{eq:loss_o}
\end{equation}
penalizing squared cosine similarities between distinct prototype pairs and forcing them to spread out as a discriminative set.

Table~\ref{tab:ablation_component_OPR_ATC}(a) verifies this design: removing the residual connection collapses global mIoU from 62.55\% to 56.93\% though tail-class pseudo-label accuracy stays high, confirming that residual aggregation protects the spatial detail required for segmentation; removing the orthogonality loss reduces global and tail pseudo-label accuracy, confirming that prototype diversity makes the global signal discriminative; and removing the suppression factor causes the largest pseudo-label accuracy drop, confirming that without semantic gating background prototypes contaminate the reconstruction.

\begin{table}[t]
  \centering
  \caption{Component-wise pseudo-label analysis on global, head, and tail classes.}
  \label{tab:pseudo_label_quality}
  \small
  \setlength{\tabcolsep}{5.5pt}
  \renewcommand{\arraystretch}{1.12}

  \newcommand{\GTcell}[1]{\makebox[0.92cm][c]{#1}}
  \newcommand{\GThead}[1]{\makebox[0.92cm][c]{\textbf{#1}}}

  \begin{tabular}{c|ccc|ccc|ccc}
    \toprule
    \multirow{2}{*}[-0.55ex]{\textbf{Method}}
    & \multicolumn{3}{c|}{\textbf{Pseudo-Label Acc (\%)}}
    & \multicolumn{3}{c|}{\textbf{Filtered-Rate (\%)}}
    & \multicolumn{3}{c}{\makebox[3.15cm][c]{\textbf{Correctly Kept GT Rate (\%)}}} \\
    \cmidrule(lr){2-4}
    \cmidrule(lr){5-7}
    \cmidrule(lr){8-10}
    & \textbf{Global} & \textbf{Head} & \textbf{Tail}
    & \textbf{Global} & \textbf{Head} & \textbf{Tail}
    & \GThead{Global} & \GThead{Head} & \GThead{Tail} \\
    \midrule
    Baseline 
    & 67.03 & 73.06 & 56.56 
    & 13.15 & 9.50 & 17.83 
    & \GTcell{58.22} & \GTcell{67.04} & \GTcell{46.47} \\

    + OPR 
    & 72.33 & 74.08 & 87.80 
    & 9.04 & 7.04 & 11.24 
    & \GTcell{63.98} & \GTcell{71.34} & \GTcell{77.93} \\

    + ATC 
    & 69.73 & 79.60 & 88.65 
    & 11.63 & 10.38 & 10.39 
    & \GTcell{64.80} & \GTcell{68.13} & \GTcell{79.44} \\

    FTC-Seg (ours) 
    & \textbf{73.83} & \textbf{84.13} & \textbf{89.03} 
    & \textbf{8.26} & \textbf{6.75} & \textbf{8.43} 
    & \GTcell{\textbf{67.15}} & \GTcell{\textbf{78.21}} & \GTcell{\textbf{80.61}} \\
    \bottomrule
  \end{tabular}
\end{table}

\begin{table}[t]
  \centering
  \caption{Ablation studies of OPR layers and FTC-Seg components.}
  \label{tab:ablation_side_by_side}
  \scriptsize
  \renewcommand{\arraystretch}{1.08}

  \begin{minipage}[t]{0.56\linewidth}
    \centering
    \textbf{(a) Effect of applying OPR to multiple encoder stages.}
    \vspace{2pt}

    \setlength{\tabcolsep}{3pt}
    \resizebox{\linewidth}{!}{%
    \begin{tabular}{c|c|c|c|c|c}
      \toprule
      \textbf{OPR Layers} &
      \textbf{mIoU} &
      \makecell{\textbf{Params}\\\textbf{(M)}} &
      \makecell{\textbf{FLOPs}\\\textbf{(G)}} &
      \makecell{\textbf{Infer.}\\\textbf{Time (ms)}} &
      \makecell{\textbf{Peak Mem.}\\\textbf{(MB)}} \\
      \midrule
      0 (Baseline) & 58.75 & 24.79 & 56.88 & 12.97 & 14279 \\
      1            & 61.92 & 24.80 & 56.90 & 13.15 & 14348 \\
      2            & \textbf{62.55} & 24.81 & 56.92 & 13.27 & 14398 \\
      3            & 61.47 & 24.82 & 56.93 & 13.43 & 14449 \\
      4            & 60.98 & 24.84 & 56.95 & 13.58 & 14497 \\
      \bottomrule
    \end{tabular}%
    }
  \end{minipage}%
  \hfill
  \begin{minipage}[t]{0.41\linewidth}
    \centering
    \textbf{(b) Components ablation analysis of FTC-Seg.}
    \vspace{2pt}

    \raisebox{-0.29in}{%
    \setlength{\tabcolsep}{5pt}
    \resizebox{\linewidth}{0.575in}{%
    \begin{tabular}{c|c|c|c|c}
      \toprule
      \textbf{Model} & \textbf{EMA} & \textbf{OPR} & \textbf{ATC} & \textbf{mIoU} \\
      \midrule
      M1 & --           & --           & --           & 35.61 \\
      M2 & $\checkmark$ & --           & --           & 58.75 \\
      M3 & $\checkmark$ & $\checkmark$ & --           & 62.55 \\
      M4 & $\checkmark$ & --           & $\checkmark$ & 61.90 \\
      M5 & $\checkmark$ & $\checkmark$ & $\checkmark$ & \textbf{63.20} \\
      \bottomrule
    \end{tabular}%
    }%
    }
  \end{minipage}
\end{table}

\paragraph{Adaptive Threshold Calibration (ATC).}
At the threshold level, the degradation cycle manifests as exclusion: once feature-level entanglement suppresses tail-class confidence, a class-agnostic fixed threshold rejects these predictions wholesale, denying tail classes any supervision and locking the cycle. A threshold-level intervention must replace the global threshold with class-specific ones that respond to each class's learning state. Self-referential adaptive thresholds estimate this state solely from the model's confidence on unlabeled data, but such a signal cannot distinguish learning progress from over-prediction bias. We argue that the intervention must combine two signals: \textbf{(i)} an internal signal reflecting how well the model has learned each class, measured against ground truth on labeled data; \textbf{(ii)} an external signal comparing the model's predicted class distribution on unlabeled data against the class prior on labeled data, exposing over or under-prediction bias that the internal signal cannot detect. ATC realizes this dual-signal design.

\begin{table}[t]
  \centering
  \caption{Dominant-class responses of 16 OPR prototypes on FSSG with 2\% labels.
BL/CG/SF/CC/SP/PO/SQ/TI/UR/DV denote Block/Circle Cage/Steel Frame/Concrete Column/Steel Plate/Pot/SquareCage/Tire/Underwater Robot/Diver.}
  \label{tab:prototype_semantics}

  \scriptsize
  \setlength{\tabcolsep}{3pt}
  \renewcommand{\arraystretch}{1.15}

  \begin{tabular*}{\linewidth}{@{\extracolsep{\fill}}llll@{}}
    \toprule
    \textbf{Prototypes 1--4} &
    \textbf{Prototypes 5--8} &
    \textbf{Prototypes 9--12} &
    \textbf{Prototypes 13--16} \\
    \midrule

    \makecell[l]{
      $\mu_{1}$: SP$\to$SP$\to$TI \\
      $P{=}14.8,\ F{=}63.6,\ H{=}99.9$
    } &
    \makecell[l]{
      $\mu_{5}$: CC$\to$CC$\to$DV \\
      $P{=}16.1,\ F{=}39.8,\ H{=}99.6$
    } &
    \makecell[l]{
      $\mu_{9}$: CC$\to$UR$\to$CC \\
      $P{=}13.6,\ F{=}56.4,\ H{=}99.8$
    } &
    \makecell[l]{
      $\mu_{13}$: SP$\to$SP$\to$CG \\
      $P{=}15.3,\ F{=}53.2,\ H{=}99.8$
    } \\
    \addlinespace[3pt]

    \makecell[l]{
      $\mu_{2}$: CC$\to$DV$\to$BL \\
      $P{=}13.2,\ F{=}53.6,\ H{=}99.8$
    } &
    \makecell[l]{
      $\mu_{6}$: SF$\to$SF$\to$UR \\
      $P{=}18.0,\ F{=}79.2,\ H{=}99.7$
    } &
    \makecell[l]{
      $\mu_{10}$: SF$\to$SF$\to$PO \\
      $P{=}16.5,\ F{=}33.9,\ H{=}99.4$
    } &
    \makecell[l]{
      $\mu_{14}$: PO$\to$PO$\to$CC \\
      $P{=}17.1,\ F{=}60.8,\ H{=}99.7$
    } \\
    \addlinespace[3pt]

    \makecell[l]{
      $\mu_{3}$: SQ$\to$UR$\to$DV \\
      $P{=}14.9,\ F{=}29.7,\ H{=}99.2$
    } &
    \makecell[l]{
      $\mu_{7}$: PO$\to$PO$\to$DV \\
      $P{=}15.4,\ F{=}43.0,\ H{=}99.6$
    } &
    \makecell[l]{
      $\mu_{11}$: DV$\to$DV$\to$PO \\
      $P{=}14.0,\ F{=}34.2,\ H{=}99.8$
    } &
    \makecell[l]{
      $\mu_{15}$: SF$\to$TI$\to$CC \\
      $P{=}12.7,\ F{=}51.6,\ H{=}99.9$
    } \\
    \addlinespace[3pt]

    \makecell[l]{
      $\mu_{4}$: CC$\to$BL$\to$CC \\
      $P{=}17.6,\ F{=}66.4,\ H{=}99.6$
    } &
    \makecell[l]{
      $\mu_{8}$: PO$\to$PO$\to$SP \\
      $P{=}16.4,\ F{=}76.4,\ H{=}99.7$
    } &
    \makecell[l]{
      $\mu_{12}$: SQ$\to$SQ$\to$PO \\
      $P{=}14.4,\ F{=}37.7,\ H{=}99.4$
    } &
    \makecell[l]{
      $\mu_{16}$: TI$\to$TI$\to$PO \\
      $P{=}13.5,\ F{=}54.3,\ H{=}99.9$
    } \\

    \bottomrule
  \end{tabular*}
\end{table}

\begin{figure}[t]
    \centering

    \begin{subfigure}{\linewidth}
        \centering
        \includegraphics[width=\linewidth]{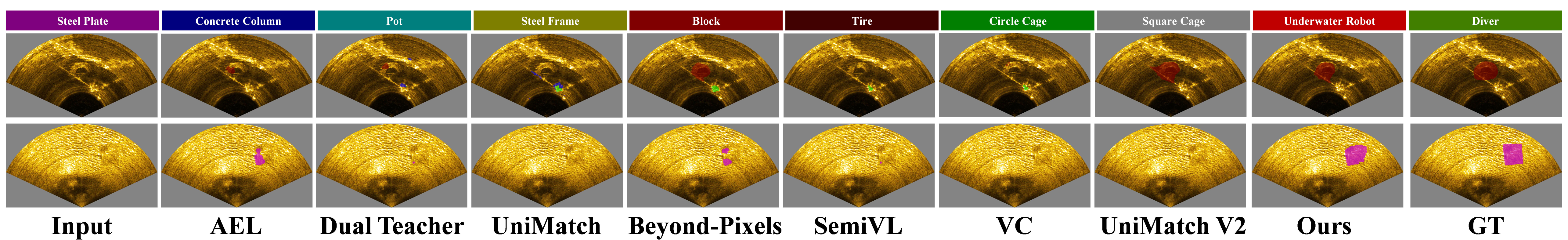}
        \vspace{-15pt}
        \caption{Qualitative segmentation results on the FSSG dataset with 2\% labeled data.}
        \label{fig:OurSonar02}
    \end{subfigure}

    \vspace{0.1em}
    
    \begin{subfigure}{\linewidth}
        \centering
        \includegraphics[width=\linewidth]{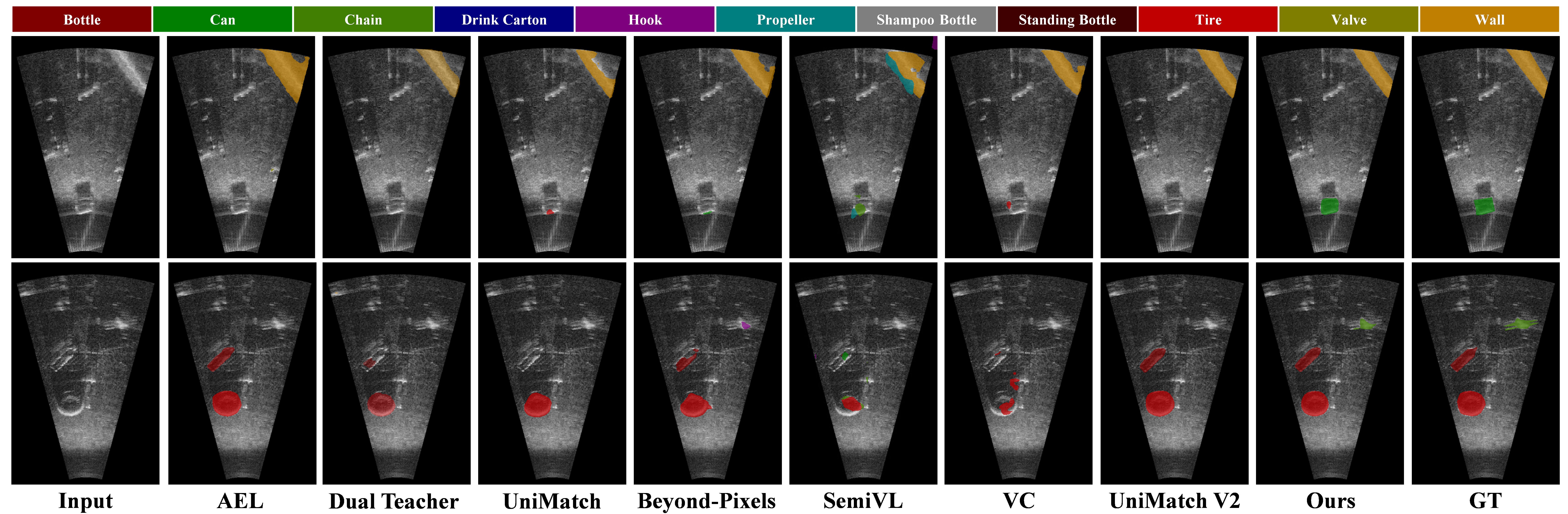}
        \vspace{-15pt}
        \caption{Qualitative segmentation results on the FLSMD dataset with 2\% labeled data.}
        \label{fig:FLSMD02-VISUAL}
    \end{subfigure}

    \caption{Qualitative segmentation results compared with state-of-the-art methods on the FSSG and FLSMD datasets with 2\% labeled data.}
    \label{fig:sonar_visual_comparison}
\end{figure}

As shown in Figure~\ref{fig:workflow}(c), we maintain three global statistics for each class $c$: the ground-truth class distribution $P_c^{l}$ and per-class learning accuracy $A_c^{l}$ on labeled data, the predicted class distribution $P_c^{u}$ on unlabeled data. To stabilize these statistics against mini-batch fluctuations—particularly severe for tail classes with small per-batch counts—we update them with EMA:
\begin{align}
P_c^{l} &\leftarrow \eta_m P_c^{l} + (1-\eta_m)\tilde{P}_c^{l},
\label{eq:pl_update} \\
P_c^{u} &\leftarrow \eta_m P_c^{u} + (1-\eta_m)\tilde{P}_c^{u},
\label{eq:pu_update} \\
A_c^{l} &\leftarrow \eta_m A_c^{l} + (1-\eta_m)\tilde{A}_c^{l},
\label{eq:al_update}
\end{align}
where $\tilde{P}_c^{l}$, $\tilde{P}_c^{u}$ and $\tilde{A}_c^{l}$ are the current-batch statistics, and $\eta_m$ is the EMA momentum (set to 0.999).

The internal signal is realized by the difficulty factor, defined as the gap between perfect accuracy and the model's actual accuracy on labeled data:
\begin{equation}
\mathcal{M}_{\text{diff}}(c) = 1 - A_c^l,
\label{eq:modulator_diff}
\end{equation}
so that a poorly-performing class is identified as difficult and granted a lower threshold to admit more of its low-confidence pseudo-labels. The external signal is realized by the bias factor, contrasting the predicted class distribution on unlabeled data against the true class prior on labeled data:
\begin{equation}
\mathcal{M}_{\text{bias}}(c) = \frac{P_c^u}{P_c^l + \epsilon} - 1,
\label{eq:modulator_bias}
\end{equation}

where $\epsilon$ is a small constant for numerical stability. A positive $\mathcal{M}_{\text{bias}}(c)$ indicates over-prediction; a negative value indicates under-prediction. This is the signal that self-referential thresholds lack: without the labeled prior $P_c^{l}$ as an anchor, a class can be heavily over-predicted yet still appear confident to the model itself. The two signals are linearly combined with the basic threshold $\tau_0$ to form the class-specific dynamic threshold:
\begin{equation}
\tau_c^* = \tau_0 - \lambda_D \cdot \mathcal{M}_{\text{diff}}(c) + \lambda_B \cdot \mathcal{M}_{\text{bias}}(c).
\label{eq:threshold_update}
\end{equation}
where $\lambda_D$ and $\lambda_B$ control the influence of each signal. The signs are deliberate: difficult classes (large $\mathcal{M}_{\text{diff}}$) lower the threshold to admit more pseudo-labels for under-learned classes, while over-predicted classes (positive $\mathcal{M}_{\text{bias}}$) raise the threshold to suppress false positives that would otherwise compound confirmation bias.

Table~\ref{tab:ablation_component_OPR_ATC}(b) verifies that the two signals identify different sets of classes to rescue. Adding $\mathcal{M}_{\text{diff}}$ alone raises tail Recall to 77.11\% by lowering thresholds for poorly-learned classes; adding $\mathcal{M}_{\text{bias}}$ alone raises Recall to 76.29\%, but by lowering thresholds for classes under-predicted relative to the labeled prior—two criteria that overlap but do not coincide. When both signals are active, Recall jumps to 86.40\%, far above either individual gain, because classes flagged by only one signal are now rescued as well. Precision rises from 77.62\% to 84.00\%, and tail IoU climbs from 49.97\% to 73.87\%, showing that the two signals expose different facets of threshold-level failure—learning-state weakness and distribution-prior bias—rather than redundantly strengthening the same effect.

\begin{figure}[t]
    \centering

    \begin{subfigure}{\linewidth}
        \centering
        \includegraphics[width=\linewidth]{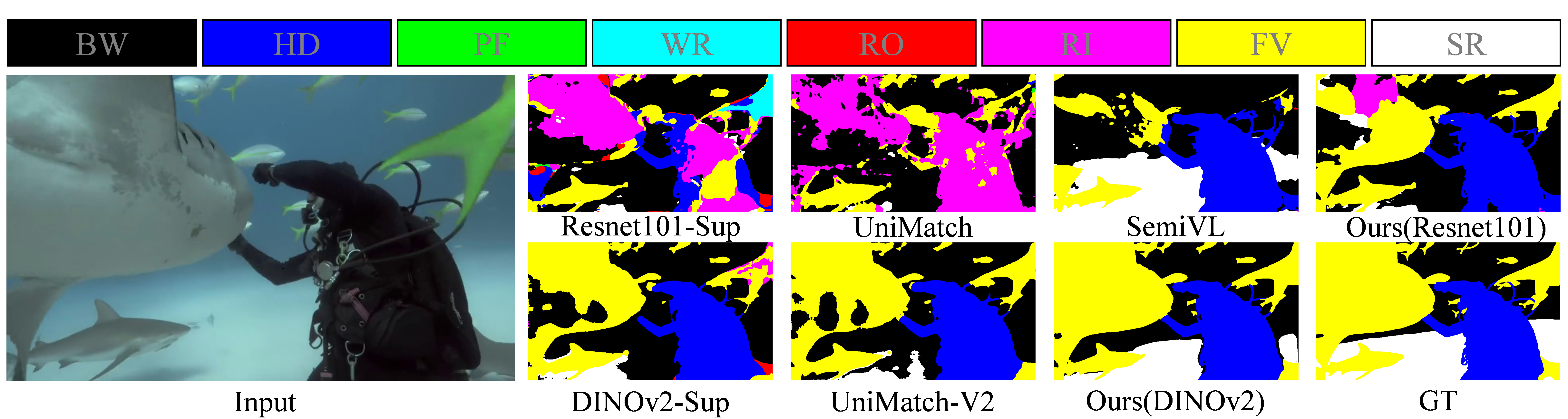}
        \vspace{-15pt}
        \caption{Qualitative segmentation results on the SUIM dataset with 2\% labeled data.}
        \label{fig:SUIM02}
    \end{subfigure}

    \vspace{0.1em}
    
    \begin{subfigure}{\linewidth}
        \centering
        \includegraphics[width=\linewidth]{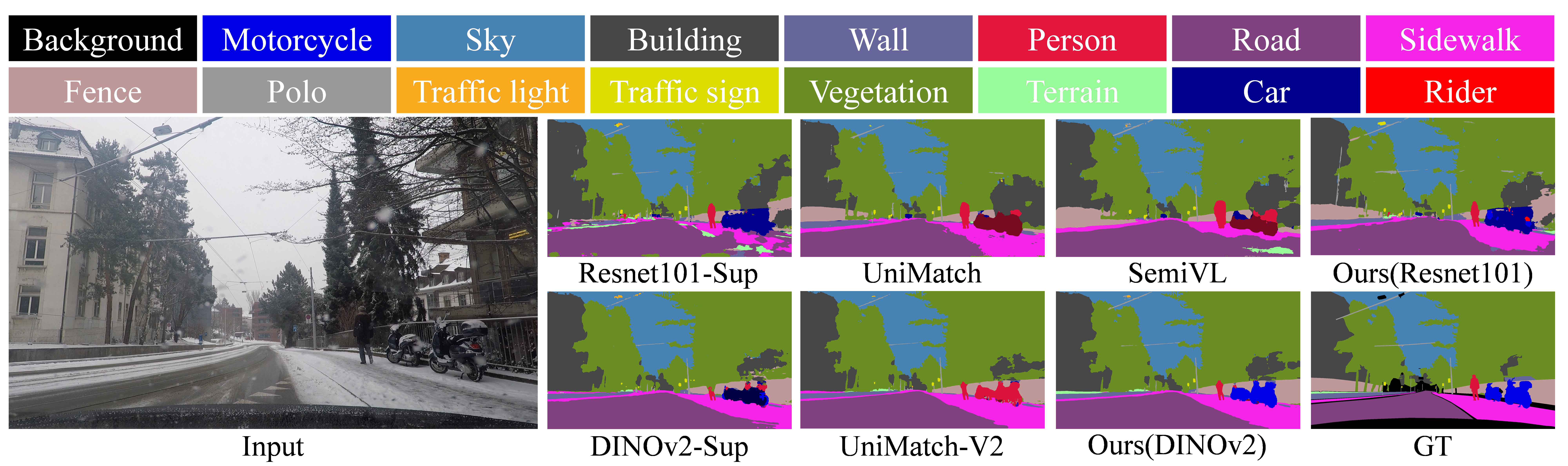}
        \vspace{-15pt}
        \caption{Qualitative segmentation results on the ACDC dataset with 2\% labeled data.}
        \label{fig:ACDC02}
    \end{subfigure}

    \caption{Qualitative segmentation results compared with state-of-the-art methods on the SUIM and ACDC datasets with 2\% labeled data.}
    \label{fig:SUIM_ACDC_Visual}
\end{figure}

\section{Experiments}
\paragraph{Experimental Setup.}
We evaluate FTC-Seg on four public benchmarks spanning three imaging-noise modalities: FLSMD~\citep{singh2021marine} and FSSG~\citep{guo2026ctfs} for acoustic speckle, SUIM~\citep{islam2020semantic} for underwater optical scattering, and ACDC~\citep{sakaridis2021acdc} for adverse-weather urban driving. All exhibit long-tailed class distributions: Standing-bottle in FLSMD, Diver in FSSG, Robots (RO) in SUIM, and Rider in ACDC account for only 1.67\%, 2.04\%, 0.52\%, and 0.02\% of pixels, respectively, supporting evaluation under combined noise and class imbalance. We sample 2\%, 5\%, and 10\% labeled data, ensuring at least one labeled instance per class, and treat the remainder as unlabeled. We use ResNet-101~\citep{he2016deep} and DINOv2-S~\citep{oquab2023dinov2} encoders for fair comparison with prior work, reporting mean Intersection over Union (mIoU) over three random seeds. All models train for 240 epochs with AdamW~\citep{loshchilov2017decoupled} (lr $5\times10^{-6}$, weight decay 0.01, polynomial decay power 0.9) on one NVIDIA RTX 3090. We compare against AEL~\citep{hu2021semi}, Dual Teacher~\citep{na2023switching}, UniMatch~\citep{yang2023revisiting}, Beyond-Pixels~\citep{howlader2024beyond}, SemiVL~\citep{hoyer2024semivl}, VC~\citep{chen2024virtual}, and UniMatch V2~\citep{yang2025unimatch}, reproduced from public code under identical settings.

\paragraph{Main Results.}
Tables~\ref{tab:FSSG_FLSMD_COMP} and \ref{tab:suim_acdc_results} report the comparison on FLSMD/FSSG and SUIM/ACDC respectively. FTC-Seg achieves the best or highly competitive mIoU in most configurations, with the most consistent advantages under low-label and tail-class regimes. Under the most challenging 2\% setting, FTC-Seg with DINOv2-S exceeds UniMatch V2 by 5.31\% mIoU on FLSMD, 4.42\% on FSSG, 8.54\% on SUIM, and 2.09\% on ACDC. This lead being preserved across three completely different noise modalities indicates that FTC-Seg corrects a failure mode generic to noisy and long-tailed pseudo-labeling, rather than an artifact tuned to any single domain.

The advantage is pronounced on long-tailed classes. Figure~\ref{fig:TailMiou} reports IoU on the rarest or second-rarest class of each dataset: Standing-bottle in FLSMD, Diver in FSSG, Robots in SUIM, and motorcycle in ACDC—we use the second-rarest for ACDC because more than half of the methods produce near-zero IoU on Rider. FTC-Seg attains the highest IoU on every class, with leads of 8.91\%, 10.04\%, 13.60\%, and 9.78\% over the second-best method, confirming that the framework rescues the systematic exclusion produced by the degradation cycle wherever noise and long-tail co-occur.

Figures~\ref{fig:sonar_visual_comparison} and~\ref{fig:SUIM_ACDC_Visual} present qualitative results across all four datasets, corroborating the quantitative findings. FTC-Seg recovers the complete object extent with sharper boundaries and fewer false positives, while comparison methods fragment or miss targets hidden in noisy backgrounds.

\paragraph{Pseudo-Label Quality on Head and Tail Classes.}
Table~\ref{tab:pseudo_label_quality} provides evidence that FTC-Seg breaks the degradation cycle of Section 1. Two observations stand out. First, the baseline's filtering is unequal: 17.83\% of tail predictions are filtered out, nearly twice the 9.50\% head rate—a manifestation of the claim that fixed thresholds disproportionately exclude tail classes. Under FTC-Seg, this gap collapses to 8.43\% versus 6.75\%, showing the disproportion is not merely reduced but largely equalized. Second, the asymmetry appears in pseudo-label accuracy: the 16.50\% head–tail gap is not only closed but reversed, with tail reaching 89.03\% against 84.13\% on head.

The two modules intervene at different levels: OPR purifies target features so tail predictions become more confident, dropping the tail filtered-rate from 17.83\% to 11.24\%; ATC directly lowers the threshold for tail classes, reaching a similar 10.39\%. Each mechanism substantially relieves the systematic exclusion, and combining them further reduces the rate to 8.43\%.

The same pattern holds at the framework level. Decomposing FTC-Seg in Table~\ref{tab:ablation_side_by_side}(b), adding the EMA teacher–student framework to the labeled-only baseline (M1 → M2) already raises mIoU from 35.61\% to 58.75\%, confirming pseudo-labeling itself is the dominant source of gain when labels are scarce. On this foundation, OPR (M3) and ATC (M4) yield 62.55\% and 61.90\% respectively, and combining both (M5) reaches 63.20\%.

\paragraph{Prototype semantics.}
Table~\ref{tab:prototype_semantics} examines 16 OPR prototypes
on FSSG with 2\% labels. Arrows track each prototype's dominant
foreground class from early to middle to final training.
Final-stage statistics report class purity ($P$), foreground
selectivity ($F$), and normalized foreground spatial entropy
($H$), all in percent. Changing dominant classes and low purity
(12.7--18.0\%) indicate that prototypes are not tied to fixed
class identities. High spatial entropy (99.2--99.9\%) and
varying foreground selectivity (29.7--79.2\%) suggest spatially
distributed responses with different foreground preferences,
consistent with class-agnostic feature reconstruction.

\paragraph{Computational Efficiency.}
Table~\ref{tab:ablation_side_by_side}(a) varies how many of the encoder's deepest stages are equipped with OPR: 0 means no OPR, 1 applies OPR only to the deepest stage, and 2 corresponds to our default architecture, where OPR is applied to the two deepest encoder stages. Higher values progressively extend OPR to shallower stages. Across all settings, parameters, FLOPs, inference time, and peak memory stay essentially flat—deviations from the no-OPR baseline are below 0.05M, 0.07G, 0.61ms, and 218MB respectively—so OPR is practically free to deploy regardless of placement. Performance peaks at two stages and degrades when OPR is extended further into shallower layers. The reason traces back to the suppression factor in Eq.~\ref{eq:Gamma}, which relies on the segmentation head's weights trained on the deepest features: prototypes from shallow features carry weaker class semantics, so projecting them onto the segmentation head yields unreliable foreground assignments. Background prototypes are then mistaken for foreground and aggregated into the reconstruction, actively injecting the interference OPR is meant to remove.

A sensitivity analysis of $ \lambda_D$, $ \lambda_B$, and $K$ on FLSMD with 2\% labeled data is provided in the appendix (Figure~\ref{fig:Parameter}); mIoU stays within 4 points of the optimum across all swept ranges, and each hyperparameter exhibits a clear single peak at the chosen default $ \lambda_D = 0.1$, $ \lambda_B = 0.05$, $K=16$.

\section{Conclusion}
This paper identified a failure mode that emerges when imaging noise and long-tailed class distributions interact within the pseudo-labeling pipeline of semantic segmentation: a vicious cycle in which noise suppresses tail-class confidence, fixed thresholds turn that suppression into systematic exclusion, and the resulting absence of supervision lets tail features keep degrading. We argued that breaking this cycle requires intervention at both the feature level and the threshold level, and instantiated this principle as FTC-Seg, in which Orthogonal Prototype Reconstruction purifies pixel features against noise-induced entanglement and Adaptive Threshold Calibration replaces the fixed threshold with class-specific ones driven by an internal learning signal and an external distribution signal. Across four benchmarks spanning acoustic speckle, underwater optical scattering and adverse-weather imaging, FTC-Seg consistently outperforms state-of-the-art methods in overall mIoU, with the largest absolute gains concentrated on the rarest tail classes of each dataset, and the most substantial improvements in pseudo-label quality also occurring on tail classes.

\begin{ack}
We thank the Graduate School of Information, Production and Systems,
Waseda University, for providing GPU computing resources.
This work received no external financial support.
The authors declare no competing interests.
\end{ack}

\bibliographystyle{plainnat}
\bibliography{Reference}

@inproceedings{yang2023revisiting,
  title={Revisiting weak-to-strong consistency in semi-supervised semantic segmentation},
  author={Yang, Lihe and Qi, Lei and Feng, Litong and Zhang, Wayne and Shi, Yinghuan},
  booktitle={Proceedings of the IEEE/CVF conference on computer vision and pattern recognition},
  pages={7236--7246},
  year={2023}
}

@inproceedings{hoyer2024semivl,
  title={SemiVL: Semi-supervised semantic segmentation with vision-language guidance},
  author={Hoyer, Lukas and Tan, David Joseph and Naeem, Muhammad Ferjad and Van Gool, Luc and Tombari, Federico},
  booktitle={European Conference on Computer Vision},
  pages={257--275},
  year={2024},
  organization={Springer}
}

@article{sohn2020fixmatch,
  title={Fixmatch: Simplifying semi-supervised learning with consistency and confidence},
  author={Sohn, Kihyuk and Berthelot, David and Carlini, Nicholas and Zhang, Zizhao and Zhang, Han and Raffel, Colin A and Cubuk, Ekin Dogus and Kurakin, Alexey and Li, Chun-Liang},
  journal={Advances in neural information processing systems},
  volume={33},
  pages={596--608},
  year={2020}
}

@article{everingham2010pascal,
  title={The pascal visual object classes (voc) challenge},
  author={Everingham, Mark and Van Gool, Luc and Williams, Christopher KI and Winn, John and Zisserman, Andrew},
  journal={International journal of computer vision},
  volume={88},
  number={2},
  pages={303--338},
  year={2010},
  publisher={Springer}
}

@inproceedings{cordts2016cityscapes,
  title={The cityscapes dataset for semantic urban scene understanding},
  author={Cordts, Marius and Omran, Mohamed and Ramos, Sebastian and Rehfeld, Timo and Enzweiler, Markus and Benenson, Rodrigo and Franke, Uwe and Roth, Stefan and Schiele, Bernt},
  booktitle={Proceedings of the IEEE conference on computer vision and pattern recognition},
  pages={3213--3223},
  year={2016}
}

@inproceedings{wei2021crest,
  title={Crest: A class-rebalancing self-training framework for imbalanced semi-supervised learning},
  author={Wei, Chen and Sohn, Kihyuk and Mellina, Clayton and Yuille, Alan and Yang, Fan},
  booktitle={Proceedings of the IEEE/CVF conference on computer vision and pattern recognition},
  pages={10857--10866},
  year={2021}
}

@inproceedings{oh2022daso,
  title={Daso: Distribution-aware semantics-oriented pseudo-label for imbalanced semi-supervised learning},
  author={Oh, Youngtaek and Kim, Dong-Jin and Kweon, In So},
  booktitle={Proceedings of the IEEE/CVF conference on computer vision and pattern recognition},
  pages={9786--9796},
  year={2022}
}

@article{wang2022freematch,
  title={Freematch: Self-adaptive thresholding for semi-supervised learning},
  author={Wang, Yidong and Chen, Hao and Heng, Qiang and Hou, Wenxin and Fan, Yue and Wu, Zhen and Wang, Jindong and Savvides, Marios and Shinozaki, Takahiro and Raj, Bhiksha and others},
  journal={arXiv preprint arXiv:2205.07246},
  year={2022}
}

@article{zhang2025noise,
  title={Noise-robust consistency regularization for semi-supervised semantic segmentation},
  author={Zhang, HaiKuan and Li, Haitao and Zhang, Xiufeng and Yang, Guanyu and Li, Atao and Du, Weisheng and Xue, Shanshan and Liu, Chi},
  journal={Neural Networks},
  volume={184},
  pages={107041},
  year={2025},
  publisher={Elsevier}
}

@article{scherer2022pseudo,
  title={Pseudo-label noise suppression techniques for semi-supervised semantic segmentation},
  author={Scherer, Sebastian and Sch{\"o}n, Robin and Lienhart, Rainer},
  journal={arXiv preprint arXiv:2210.10426},
  year={2022}
}

@article{tarvainen2017mean,
  title={Mean teachers are better role models: Weight-averaged consistency targets improve semi-supervised deep learning results},
  author={Tarvainen, Antti and Valpola, Harri},
  journal={Advances in neural information processing systems},
  volume={30},
  year={2017}
}

@inproceedings{chen2021semi,
  title={Semi-supervised semantic segmentation with cross pseudo supervision},
  author={Chen, Xiaokang and Yuan, Yuhui and Zeng, Gang and Wang, Jingdong},
  booktitle={Proceedings of the IEEE/CVF conference on computer vision and pattern recognition},
  pages={2613--2622},
  year={2021}
}

@article{yang2025unimatch,
  title={Unimatch v2: Pushing the limit of semi-supervised semantic segmentation},
  author={Yang, Lihe and Zhao, Zhen and Zhao, Hengshuang},
  journal={IEEE Transactions on Pattern Analysis and Machine Intelligence},
  volume={47},
  number={4},
  pages={3031--3048},
  year={2025},
  publisher={IEEE}
}

@inproceedings{howlader2024beyond,
  title={Beyond pixels: Semi-supervised semantic segmentation with a multi-scale patch-based multi-label classifier},
  author={Howlader, Prantik and Das, Srijan and Le, Hieu and Samaras, Dimitris},
  booktitle={European Conference on Computer Vision},
  pages={342--360},
  year={2024},
  organization={Springer}
}

@article{zhang2021flexmatch,
  title={Flexmatch: Boosting semi-supervised learning with curriculum pseudo labeling},
  author={Zhang, Bowen and Wang, Yidong and Hou, Wenxin and Wu, Hao and Wang, Jindong and Okumura, Manabu and Shinozaki, Takahiro},
  journal={Advances in neural information processing systems},
  volume={34},
  pages={18408--18419},
  year={2021}
}

@article{hu2021semi,
  title={Semi-supervised semantic segmentation via adaptive equalization learning},
  author={Hu, Hanzhe and Wei, Fangyun and Hu, Han and Ye, Qiwei and Cui, Jinshi and Wang, Liwei},
  journal={Advances in Neural Information Processing Systems},
  volume={34},
  pages={22106--22118},
  year={2021}
}

@article{na2023switching,
  title={Switching temporary teachers for semi-supervised semantic segmentation},
  author={Na, Jaemin and Ha, Jung-Woo and Chang, Hyung Jin and Han, Dongyoon and Hwang, Wonjun},
  journal={Advances in Neural Information Processing Systems},
  volume={36},
  pages={40367--40380},
  year={2023}
}

@article{chen2024virtual,
  title={Virtual category learning: A semi-supervised learning method for dense prediction with extremely limited labels},
  author={Chen, Changrui and Han, Jungong and Debattista, Kurt},
  journal={IEEE transactions on pattern analysis and machine intelligence},
  volume={46},
  number={8},
  pages={5595--5611},
  year={2024},
  publisher={IEEE}
}

@article{guo2026ctfs,
  title={CTFS: Collaborative Teacher Framework for Forward-Looking Sonar Image Semantic Segmentation with Extremely Limited Labels},
  author={Guo, Ping and Li, Chengzhou and Meng, Guanchen and Jia, Qi and Liu, Jinyuan and Liu, Zhu and Liu, Yu and Luo, Zhongxuan and Fan, Xin},
  journal={arXiv preprint arXiv:2603.21071},
  year={2026}
}

@inproceedings{singh2021marine,
  title={The marine debris dataset for forward-looking sonar semantic segmentation},
  author={Singh, Deepak and Valdenegro-Toro, Matias},
  booktitle={Proceedings of the ieee/cvf international conference on computer vision},
  pages={3741--3749},
  year={2021}
}

@inproceedings{islam2020semantic,
  title={Semantic segmentation of underwater imagery: Dataset and benchmark},
  author={Islam, Md Jahidul and Edge, Chelsey and Xiao, Yuyang and Luo, Peigen and Mehtaz, Muntaqim and Morse, Christopher and Enan, Sadman Sakib and Sattar, Junaed},
  booktitle={2020 IEEE/RSJ international conference on intelligent robots and systems (IROS)},
  pages={1769--1776},
  year={2020},
  organization={IEEE}
}

@inproceedings{sakaridis2021acdc,
  title={ACDC: The adverse conditions dataset with correspondences for semantic driving scene understanding},
  author={Sakaridis, Christos and Dai, Dengxin and Van Gool, Luc},
  booktitle={Proceedings of the IEEE/CVF international conference on computer vision},
  pages={10765--10775},
  year={2021}
}

@inproceedings{he2016deep,
  title={Deep residual learning for image recognition},
  author={He, Kaiming and Zhang, Xiangyu and Ren, Shaoqing and Sun, Jian},
  booktitle={Proceedings of the IEEE conference on computer vision and pattern recognition},
  pages={770--778},
  year={2016}
}

@article{oquab2023dinov2,
  title={Dinov2: Learning robust visual features without supervision},
  author={Oquab, Maxime and Darcet, Timoth{\'e}e and Moutakanni, Th{\'e}o and Vo, Huy and Szafraniec, Marc and Khalidov, Vasil and Fernandez, Pierre and Haziza, Daniel and Massa, Francisco and El-Nouby, Alaaeldin and others},
  journal={arXiv preprint arXiv:2304.07193},
  year={2023}
}

@article{loshchilov2017decoupled,
  title={Decoupled weight decay regularization},
  author={Loshchilov, Ilya and Hutter, Frank},
  journal={arXiv preprint arXiv:1711.05101},
  year={2017}
}

@inproceedings{li2026rsod,
  title={Rsod: Reliability-guided sonar image object detection with extremely limited labels},
  author={Li, Chengzhou and Guo, Ping and Meng, Guanchen and Jia, Qi and Liu, Jinyuan and Liu, Zhu and Liu, Xiaokang and Liu, Yu and Luo, Zhongxuan and Fan, Xin},
  booktitle={Proceedings of the AAAI Conference on Artificial Intelligence},
  volume={40},
  number={8},
  pages={6055--6063},
  year={2026}
}

@inproceedings{feng2026semi,
  title={Semi-supervised Source Detection in Astronomical Images: New Benchmark and Strong Baseline},
  author={Feng, Longhan and Cao, Zihuang and Luo, Ali and Guo, Yuanhao and Yao, Shuilian and Guo, Yixin and Jia, Qi and Liu, Yu},
  booktitle={Proceedings of the 32nd ACM SIGKDD Conference on Knowledge Discovery and Data Mining V. 2},
  pages={10901--10912},
  year={2026}
}

@inproceedings{zhang2025zero,
  title={Zero-shot learning in industrial scenarios: New large-scale benchmark, challenges and baseline},
  author={Zhang, Zekai and Chen, Qinghui and Xiong, Maomao and Ding, Shijiao and Su, Zhanzhi and Yao, Xinjie and Sun, Yiming and Bai, Cong and Zhang, Jinglin},
  booktitle={Proceedings of the AAAI Conference on Artificial Intelligence},
  volume={39},
  number={10},
  pages={10357--10366},
  year={2025}
}

@inproceedings{liu2024two,
  title={Two teachers are better than one: semi-supervised elliptical object detection by dual-teacher collaborative guidance},
  author={Liu, Yu and Feng, Longhan and Jia, Qi and Liu, Zezheng and Cao, Zi-Huang},
  booktitle={Proceedings of the 32nd ACM International Conference on Multimedia},
  pages={6355--6363},
  year={2024}
}


\appendix
\renewcommand{\thefigure}{A\arabic{figure}}
\setcounter{figure}{0}

\section{Technical appendices and supplementary material}

\begin{table}[htbp]
  \centering
  \caption{Joint calibration across four datasets and three labeled-data
  ratios. M3 uses EMA+OPR, M4 uses EMA+ATC, and M5 uses EMA+OPR+ATC.
  Results are mean mIoU (\%) over three matched seeds shared by M3--M5 within each setting; these seeds differ from those used in the main comparison tables.
  $\Delta=\mathrm{M5}-\max(\mathrm{M3},\mathrm{M4})$ is measured in
  percentage points. Two-sided paired $t$-tests compare M5 against
  the stronger individual module selected by mean mIoU.
  Reported $p$-values are Holm-adjusted across the 12 settings;
  $\dagger$ denotes $p\geq0.05$.}
  \label{tab:joint_calibration_gains}
  \small
  \setlength{\tabcolsep}{7pt}
  \renewcommand{\arraystretch}{1.08}
  \begin{tabular}{llrrrrr}
    \toprule
    Dataset & Labels & M3 & M4 & M5 & $\Delta$ & Holm $p$ \\
    \midrule
    \multirow{3}{*}{FLSMD}
      & 2\%  & 59.80 & 60.17 & \textbf{63.07} & +2.90 & 0.0436 \\
      & 5\%  & 66.72 & 66.49 & \textbf{68.84} & +2.12 & 0.0225 \\
      & 10\% & 71.24 & 70.99 & \textbf{73.49} & +2.25 & 0.0356 \\
    \midrule
    \multirow{3}{*}{FSSG}
      & 2\%  & 60.87 & 61.24 & \textbf{63.20} & +1.96 & 0.0448 \\
      & 5\%  & 62.73 & 62.46 & \textbf{66.31} & +3.58 & 0.0160 \\
      & 10\% & 63.97 & 64.16 & \textbf{68.01} & +3.85 & 0.0086 \\
    \midrule
    \multirow{3}{*}{SUIM}
      & 2\%  & 64.49 & 63.79 & \textbf{67.04} & +2.55 & 0.0192 \\
      & 5\%  & 67.14 & 67.33 & \textbf{70.33} & +3.00 & 0.0041 \\
      & 10\% & 67.53 & 68.17 & \textbf{70.49} & +2.32 & 0.0086 \\
    \midrule
    \multirow{3}{*}{ACDC}
      & 2\%  & 58.32 & 59.07 & \textbf{61.03} & +1.96
      & $0.0769^{\dagger}$ \\
      & 5\%  & 64.92 & 64.12 & \textbf{68.29} & +3.37 & 0.0448 \\
      & 10\% & 67.37 & 68.07 & \textbf{71.60} & +3.53 & 0.0356 \\
    \bottomrule
  \end{tabular}
\end{table}

\begin{table}[htbp]
  \centering
  \caption{Prototype formulation comparison on FSSG with 2\% labels
  using the same seed. Original OPR uses 16 learnable prototypes
  without predefined class identities. Class-anchored OPR uses one
  prototype per class, initialized from labeled class-mean features
  and assigned a fixed class identity. Tail mIoU measures segmentation
  on rare classes; pseudo-label mIoU evaluates accepted pseudo-labels
  against ground truth on unlabeled training pixels. All metrics
  are percentages.}
  \label{tab:class_anchored_opr}
  \small
  \setlength{\tabcolsep}{8pt}
  \renewcommand{\arraystretch}{1.1}
  \begin{tabular}{lrrr}
    \toprule
    Method & mIoU & Tail mIoU & Pseudo-label mIoU \\
    \midrule
    EMA baseline
      & 58.78 & 5.33 & 53.49 \\
    Original OPR
      & \textbf{62.55} & \textbf{17.19} & \textbf{73.86} \\
    Class-anchored OPR
      & 54.87 & 8.93 & 58.25 \\
    \bottomrule
  \end{tabular}
\end{table}

\begin{table}[htbp]
  \centering
  \caption{Sensitivity to misspecifying the foreground set
  $\mathcal{C}_{fg}$ on FSSG with 2\% labels.
  All variants use EMA+OPR without ATC, a fixed pseudo-label
  threshold of 0.95, and the same seed. Background and Diver have
  category IDs 0 and 10, respectively. Only the gating set changes;
  segmentation output classes remain unchanged.
  IoU values are percentages, and $\Delta$ denotes the mIoU
  change from the correct setting in percentage points.}
  \label{tab:foreground_set_sensitivity}
  \small
  \setlength{\tabcolsep}{6pt}
  \renewcommand{\arraystretch}{1.1}
  \begin{tabular}{llrrr}
    \toprule
    Setting & $\mathcal{C}_{fg}$ & mIoU & $\Delta$ & Diver IoU \\
    \midrule
    Correct
      & $\{1,\ldots,10\}$ & 62.55 & -- & 17.50 \\
    Add background
      & $\{0,\ldots,10\}$ & 59.80 & $-2.75$ & 10.11 \\
    Omit Diver
      & $\{1,\ldots,9\}$ & 58.42 & $-4.13$ & 11.49 \\
    Add background + omit Diver
      & $\{0,\ldots,9\}$ & 53.09 & $-9.46$ & 2.47 \\
    \bottomrule
  \end{tabular}
\end{table}

\begin{table}[htbp]
  \centering
  \caption{Sensitivity of the noise--long-tail interaction to
  corruption severity. Each cell reports the corruption parameter
  and interaction $I=CL-OL-CB+OB$ in percentage points.
  OB, OL, CB, and CL denote tail-class pseudo-label filtering rates
  under original-balanced, original-long-tail, corrupted-balanced,
  and corrupted-long-tail conditions, respectively.
  Positive $I$ indicates super-additive exclusion.
  Larger speckle $\sigma$, smaller scattering transmission $t$,
  and larger $a$ with smaller $b$ for fog indicate stronger corruption.
  The checkpoint, image lists, class definitions, threshold, and
  random seed are held fixed across severity levels.}
  \label{tab:corruption_severity}
  \small
  \setlength{\tabcolsep}{7pt}
  \renewcommand{\arraystretch}{1.15}
  \begin{tabular}{llll}
    \toprule
    Dataset / corruption & Mild: parameter / $I$
    & Medium: parameter / $I$ & Severe: parameter / $I$ \\
    \midrule
    FLSMD / speckle
      & $\sigma=0.15$ / $+0.89$
      & $\sigma=0.35$ / $+1.83$
      & $\sigma=0.60$ / $+2.92$ \\
    SUIM / scattering
      & $t=0.80$ / $+1.04$
      & $t=0.60$ / $+1.56$
      & $t=0.40$ / $+1.22$ \\
    ACDC / fog
      & $(a,b)=(1.5,2.0)$ / $-0.11$
      & $(a,b)=(2.5,1.7)$ / $+1.55$
      & $(a,b)=(3.0,1.4)$ / $+2.23$ \\
    \bottomrule
  \end{tabular}
\end{table}

\begin{table}[htbp]
  \centering
  \caption{Single-seed controlled evaluation on Pascal VOC using
  the public 1/16 semi-supervised split with 155 labeled and
  5,507 unlabeled training images. A long-tailed distribution
  with imbalance ratio 20 is constructed, and Gaussian noise
  with $\sigma=0.18$ is applied to all labeled and unlabeled
  training images in the noisy condition. Both conditions use
  the same training split and unchanged official validation set.
  Results are validation mIoU (\%). Noise-induced change is the
  noisy result minus the long-tail-only result; gains and changes
  are reported in percentage points.}
  \label{tab:voc_controlled}
  \small
  \setlength{\tabcolsep}{8pt}
  \renewcommand{\arraystretch}{1.1}
  \begin{tabular}{lrrr}
    \toprule
    Method & Long-tail only & Long-tail + noise & Noise-induced change \\
    \midrule
    Baseline
      & 70.76 & 63.18 & $-7.58$ \\
    FTC-Seg
      & \textbf{75.36} & \textbf{69.11} & $-6.25$ \\
    \midrule
    FTC-Seg gain
      & $+4.60$ & $+5.93$ & -- \\
    \bottomrule
  \end{tabular}
\end{table}

\begin{table}[htbp]
  \centering
  \caption{Inference cost of compared methods and individual
  modules with DINOv2-S and a DPT head. Parameters and FLOPs
  are measured for a single $1\times3\times518\times518$ input.
  FPS is measured after warm-up on an RTX 4090.
  The upper block reports method comparisons; the lower block
  reports component ablations. FSSG results with 2\% labels are
  included as performance references. ATC operates during training
  and adds no inference-time computation.}
  \label{tab:inference_cost}
  \small
  \setlength{\tabcolsep}{7pt}
  \renewcommand{\arraystretch}{1.1}
  \begin{tabular}{lrrrr}
    \toprule
    Method & Params (M) & FLOPs (G) & FPS & mIoU (\%) \\
    \midrule
    \multicolumn{5}{l}{\textit{Method comparison}} \\
    Labeled only
      & 24.79 & 56.94 & 122.62 & 35.61 \\
    UniMatch V2
      & 24.79 & 56.94 & 122.62 & 58.78 \\
    FTC-Seg
      & 24.81 & 56.96 & 120.93 & 63.20 \\
    \midrule
    \multicolumn{5}{l}{\textit{Component ablation}} \\
    M2: EMA
      & 24.79 & 56.88 & 122.62 & 58.75 \\
    M3: EMA + OPR
      & 24.81 & 56.92 & 120.93 & 62.55 \\
    M4: EMA + ATC
      & 24.79 & 56.94 & 122.62 & 61.90 \\
    M5: EMA + OPR + ATC
      & 24.81 & 56.96 & 120.93 & 63.20 \\
    \bottomrule
  \end{tabular}
\end{table}

\begin{figure}[htbp]
    \centering
    \hspace*{-0.17in}
    \includegraphics[width=\linewidth,height=0.27\linewidth]{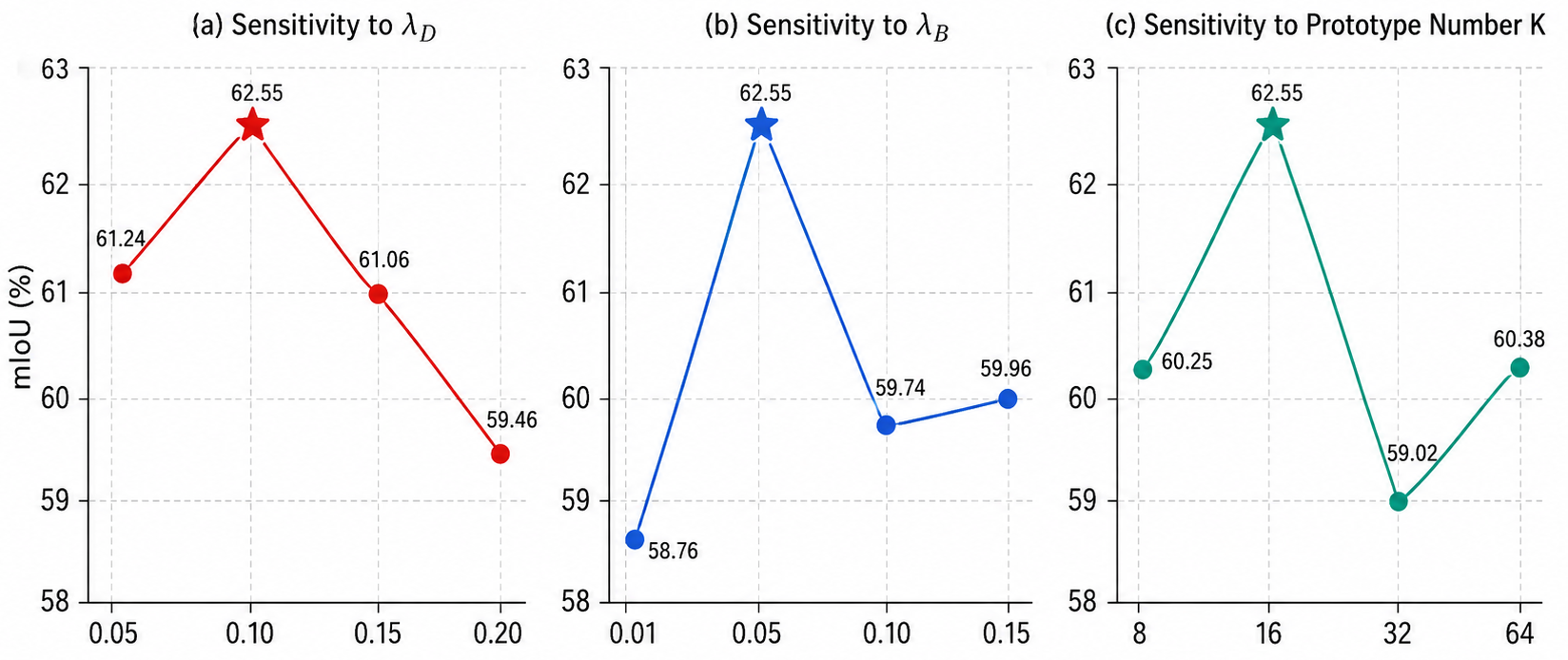}  
    \caption{Sensitivity analysis of key hyperparameters in FTC-Seg framework on the FLSMD dataset with 2\% labeled data.}
    \label{fig:Parameter}
\end{figure}
\FloatBarrier


\end{document}